\documentclass{article} 
\usepackage{times}
\usepackage{booktabs}
\usepackage{verbatim}
\usepackage{setspace}
\usepackage{graphicx}
\usepackage[centertags]{amsmath}
\usepackage{amsfonts}
\usepackage{multirow}
\usepackage{multicol}
\usepackage{amssymb}
\usepackage{amsmath}
\usepackage{amsthm}
\usepackage[noend]{algpseudocode}
\usepackage{algorithm}

\theoremstyle{plain}
\newtheorem{thm}{Theorem}
\newtheorem{cor}{Corollary}[section]
\newtheorem{lem}{Lemma}[section]

\theoremstyle{definition}
\newtheorem{defn}{Definition}[section]
\theoremstyle{remark}
\newtheorem{rem}{Remark}[section]

\title{Density Estimation via Discrepancy}

\author{
Kun Yang$^{1}$, ~Hao Su$^{2}$, ~Wing Hung Wong$^{3}$\\
$^{1}$Institute of Computational and Mathematical Engineering\\
$^{2}$Department of Computer Science\\
$^{3}$Department of Statistics, Department of Health Research and Policy\\
Stanford University
}

\begin{document}
\maketitle

\begin{abstract}
Given i.i.d samples from some unknown continuous density on hyper-rectangle $[0, 1]^d$, we attempt to learn a piecewise constant function that approximates this underlying density non-parametrically. Our density estimate is defined on a binary split of $[0, 1]^d$ and built up sequentially according to discrepancy criteria; the key ingredient is to control the discrepancy adaptively in each sub-rectangle to achieve overall bound. We prove that the estimate, even though simple as it appears, preserves most of the estimation power. By exploiting its structure, it can be directly applied to some important pattern recognition tasks such as mode seeking and density landscape exploration. We demonstrate its applicability through simulations and examples.
\end{abstract}

\section{Introduction}
\begin{sloppypar}
An important machine learning task is to efficiently summarize large-scale high-dimensional data into compact form at multiple resolutions. Since these data are typically sampled from multi-modal distributions, a natural choice would be using nonparametric density estimation methods. Classic empirical distribution (ED) and kernel density estimation (KDE) play an important role in nonparametric density estimation. Besides their long noticed drawbacks (e.g., ED is non-continuous; KDE is sensitive to the choice of bandwidth and scales poorly in high dimensions), they are not good summarization tools in dealing with data with high dimension and large size, e.g., evaluating them involves each data point and their functional forms provide little direct information of the ``landscape'' of the distribution.
\end{sloppypar}

In this paper, we consider domain partition based approach for density estimation. The use of domain partition dates back to histogram, which is still an ubiquitous tool in data analysis today; however, its non-scalability in high dimensions limits its applications. Motivated by the usefulness of histogram and the attempts to adapt it for multivariate cases, we propose a novel nonparametric density estimation method. In order to approximate distributions with continuous densities, the functional class of densities we adopt is guided by two principles: \emph{simple} and \emph{rich}. \emph{Simple} means that the functions in the class have concise forms and are cheap to evaluate; moreover, they enjoy nice structures. \emph{Rich} means that any continuous density can be approximated by functions in the class at any accuracy.

We choose \emph{piecewise constant functions supported on binary partitions} (Section 2), which is a more scalable and adaptive partitioning scheme as opposed to mesh used by histogram. Most importantly, this class satisfies both requirements: 1) functions in the class are defined by the underlying partitions of the domain and can be displayed by binary trees; consequently, trees provide a hierarchical summary of the distribution and can reveal its landscape in ``multi-resolution''; 2) it is well known in calculus that any continuous function can be approximated by piecewise constant functions.

\begin{figure}
  \center
  \includegraphics[width = .5\textwidth]{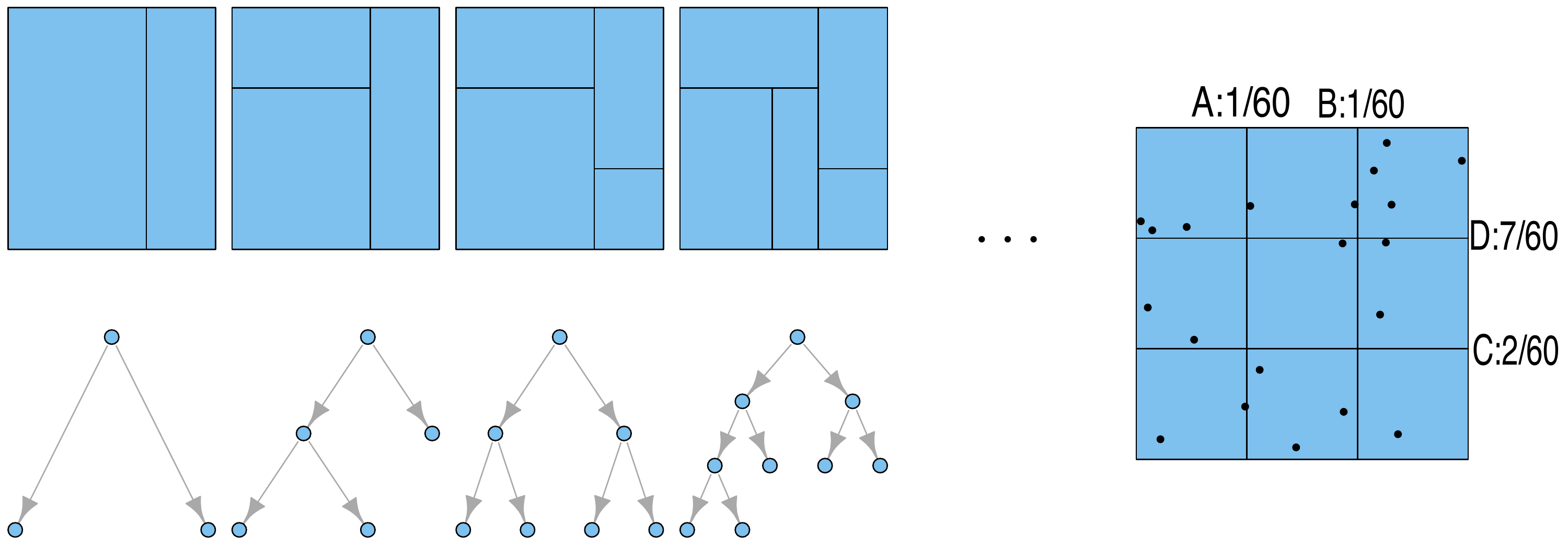}
  \caption{Left: a sequence of binary partition and the corresponding tree representation; if we encode partitioning information (e.g., the location where the split occurs) in the nodes, the mapping is one-to-one. Right: the gaps with $m = 3$, we split the rectangle at location D, which corresponds to the largest gap, if it does not satisfy \eqref{cond}.}
  \label{binary}
\end{figure}

Since the distributions conditioned on each sub-rectangle are uniform for piecewise constant densities, we construct the density estimator based on discrepancy criteria. We show that, in rather general settings, our estimated density, simple as it appears, preserves most of the estimation power, i.e., controls the integration error for the family of functions with finite total variation and finite variance, under the same convergence rate as Monte Carlo methods. Our algorithm, by exploiting the sequential build-up of binary partition as shown in Figure \ref{binary}, can find the density efficiently. It is also worth noting that the family of functions with finite total variation and finite variance is rather broad and is sufficient for practical use.

In summary, we highlight our contributions as follows:
\vspace{-2mm}
\begin{itemize}
  \item To our knowledge, this is the \emph{first} error analysis on binary partition based density estimation, which interconnects the study of Quasi-Monte Carlo and density estimation.
  \item We establish an $O(n^{-1 / 2})$ error bound of the estimator, which is \emph{optimal} in the sense of Monte Carlo integration. Simulations support the tightness.
  \item Our method is a general \emph{data summarization} tool and is readily applicable to important learning tasks such as \emph{mode seeking} and \emph{level-set tree} construction.
\end{itemize}
\vspace{-2mm}
\section{Density Estimation via Discrepancy}
Let $\Omega$ be a hyper-rectangle in $\mathbb{R}^d$. A binary partition $\mathcal{B}$ on $\Omega$ is a collection of sub-rectangles whose union is $\Omega$. Starting with $\mathcal{B}_1 = \{\Omega\}$ at level $1$ and $\mathcal{B}_t = \{\Omega_1, ..., \Omega_t\}$ at level $t$, $\mathcal{B}_{t + 1}$ is produced by dividing one of regions in $\mathcal{B}_t$ along one of its coordinates, then combining both sub-rectangles with the rest of regions in $\mathcal{B}_t$; continuing with this fashion, one can generate any binary partition at any level (Figure \ref{binary}).

At each stage of sequential built-up of binary partition, to decide whether the sub-rectangle deserves further partitioning, we need to check whether the points in it are ``relative'' uniformly scattered. Discrepancy, which is widely used in the analysis of Quasi-Monte Carlo methods, is a set of criteria to measure the uniformity of points in $[0, 1]^d$. The precise definitions of the discrepancy and the variation depend on the space of integrands. The classic star discrepancy, which is used to bound the error of Quasi-Monte Carlo integration, is defined as,
\begin{defn}
    The star discrepancy of $x_1, ..., x_n\in [0, 1]^d$ is
    \small
    \begin{equation}
    D_{n}^*(x_1, ..., x_n) = \sup_{a\in[0, 1]^d}\Big|\frac{1}{n}\sum_{i = 1}^n\mathbf{1}_{x_i\in[0, a)} - \prod_{i = 1}^da_i\Big|
    \label{DI}
    \end{equation}
    \normalsize
\end{defn}
The error bound is the famous Koksma-Hlawka inequality and the proof is included in \cite{Kuipers2012}.
\begin{thm}
    (Koksma-Hlawka inequality). Let $x_1, x_2, ..., x_n\in [0, 1]^d$ and $f$ be defined on $[0, 1]^d$, then
    \small
    \[\Big|\int_{[0, 1]^d}f(x)dx - \frac{1}{n}\sum_{i = 1}^n f(x_i)\Big|\leq D_{n}^*(x_1, ..., x_n)V_{HK}^{[0, 1]^d}(f)\]
    \normalsize
    where $s = \{1, ..., d\}$ and $V_{HK}^{[0, 1]^d}(f)$ is the total variation in the sense of Hardy and Krause, e.g., for any hyper-rectangle $[a, b]$, if all the involved partial derivatives of $f$ are continuous on $[a, b]$, then
    \small
    \begin{equation}
    V_{HK}^{[a, b]}(f) = \sum_{u\subsetneq\{1, ..., d\}}\Big\|\frac{\partial^{|u|}f}{\partial x_{u}}\Big|_{x_{s-u} = b_{s-u}}\Big\|_1^1
    \label{VHK}
    \end{equation}
    \normalsize
    \label{KH}
\end{thm}
We split the sub-rectangle when the discrepancy of points in it is larger than some threshold value. In order to find a good location to split for $[a, b] = \prod_{j = 1}^d[a_j, b_j]$, we divide $j$th dimension into $m$ bins $[a_j, a_j + (b_j - a_j) / m], ..., [a_j + (b_j - a_j) (m - 2) / m, a_j + (b_j - a_j) (m - 1) / m]$ and keep track of the gaps at $a_j + (b_j - a_j) / m, ..., a_j + (b_j - a_j) (m - 1) / m$, where the gap $g_{jk}$ is defined as $|(1/n)\sum_{i = 1}^n\mathbf{1}(x_{ij} < a_j + (b_j - a_j) k / m) - k / m|$ for $k = 1, ..., (m - 1)$, there are in total $(m - 1)d$ gaps recorded (Figure \ref{binary}). $[a, b]$ is split into two sub-rectangles along the dimension and location corresponding to maximum gap (Figure \ref{binary}). The complete algorithm is given in Algorithm \ref{algo}, and is explained in detail in the following sections.\\
\vspace{-3mm}
\begin{algorithm}[t!]
\caption{Density Estimation via Discrepancy}
Let $P(\cdot)$ define the points and $\Pr(\cdot)$ define the probability mass in a hyper-rectangle respectively. W.L.G, we assume that $\Omega = [0, 1]^d$ and $P(\Omega) = \{x_i = (x_{i1}, ..., x_{id}), x_i\in\Omega\}_{i = 1}^n$ are i.i.d samples drawn from an underlying distribution.
{\fontsize{8}{1}\selectfont
\begin{algorithmic}[1]
\Procedure{density-estimator}{$\Omega, P, m, \theta$}
\State $\mathcal{B} = \{[0, 1]^d\}$, $\Pr([0, 1]^d) = 1$
\While{true}
\State $\mathcal{B}' = {\emptyset}$
\For{each $r_i = [a_i, b_i]$ in $\mathcal{B}$}
\State Calculate gaps $\{g_{jk}\}_{j = 1, ..., d, k = 1, ..., m - 1}$
\State Scale $P(r_i) = \{x_{i_j}\}_{j = 1}^{n_i}$ to $\tilde{P} = \{\tilde{x}_{i_j} = (\frac{x_{i_j, 1} - a_{i1}}{b_{i1}}, ..., \frac{x_{i_j, d} - a_{id}}{b_{id}})\}_{j = 1}^{n_i}$
\If{$P(r_i)\neq\emptyset$ and $D_{n_i}^*(\tilde{P}) > \alpha_i D_{n_i, d}^*$}\Comment by Condition \eqref{cond} in Theorem \ref{theorem3}\\
\Comment These values can also be recorded to save computation
\State Divide $r_i$ into $r_{i1} = [a_{i1}, b_{i1}]$ and $r_{i2} = [a_{i2}, b_{i1}]$ along the max gap (Figure \ref{binary}).
\State $\Pr(r_{i1}) = \Pr(r_i)\frac{|P(r_{i1})|}{n_i}$, $\Pr(r_{i2}) = \Pr(r_i) - \Pr(r_{i1})$
\State $\mathcal{B}' = \mathcal{B}'\cup\{r_{i1}, r_{i2}\}$
\Else $\ \mathcal{B}' = \mathcal{B}'\cup\{r_i\}$
\EndIf
\EndFor
\If{$\mathcal{B}'\neq\mathcal{B}$} $\mathcal{B} = \mathcal{B}'$
\Else $\ $return $\mathcal{B}, \Pr(\cdot)$
\EndIf
\EndWhile
\EndProcedure
\end{algorithmic}
}
\begin{rem}
  Zero probability is not desirable in some applications; it can be avoided by adding pseudo count (Laplace smoother) $\alpha$ in line 11, i.e., $\Pr(r_{i1}) = \Pr(r_i)\frac{|P(r_{i1})| + \alpha}{n_i + 2\alpha}$. Density $d(r_i)$ is recovered by $\Pr(r_i) / \prod_{j = 1}^d(b_{ij} - a_{ij})$.
\end{rem}
\begin{rem}
  The binary tree shown in Figure \ref{binary} can be constructed as a byproduct and the user can specify the deepest level to terminate the algorithm.
  \label{algo_rem_2}
\end{rem}
\label{algo}
\end{algorithm}

The density, which is a piecewise constant function, is
\small
 \begin{equation}
 \hat{p}(x) = \sum_{i = 1}^l d(r_i)\mathbf{1}\{x\in r_i\}
 \label{eq1}
 \end{equation}
\normalsize
where $\mathbf{1}$ is the indicator function; $\{r_i, d(r_i)\}_{i = 1}^l$ is a list of pairs of sub-rectangles and corresponding densities. Since the number of sub-regions is far less than data size, the partition is a concise representation of the data; in Experimental Results section, we demonstrate how $\hat{p}(x)$ can be leveraged in various machine learning applications.

Compared to histogram which has the same form \eqref{eq1} but suffers from curse of dimensionality, the rational behind our adaptive partition scheme is to avoid splitting the sub-rectangle where the data are relatively uniform. One classic results of histogram \cite{Gyorfi2002} states that if for each sphere $S$ centered at the origin
\small\[h_n\rightarrow 0\text{ as } n\rightarrow 0, \lim_{n\rightarrow \infty}\frac{|\{r^n_i: r^n_i\cap S \neq \emptyset\}|}{n} = 0\]\normalsize
then
\small\[\lim_{n\rightarrow\infty}\mathbf{E}\|p(x) - \hat{p}_n(x)\|_{1} = 0, \lim_{n\rightarrow\infty}\|p(x) - \hat{p}_n(x)\|_1 = 0, a.s.\]\normalsize
where $h_n$ is the bandwidth of histogram, $\{r^n_i\}$ is the histogram bins at sample size $n$ and $\hat{p}_n(x)$ is the density estimation from $\{r^n_i\}$ by  \eqref{eq1}.

The key tool in proving its convergence is Lebesgue Density Theorem. However, our method cannot guarantee the size of each sub-rectangle shrinks to 0, which causes the technical difficulty in proving its consistency. Instead, we establish a weaker convergence result in the following section and leave the pointwise convergence as an open problem.

\section{Theoretical Results}
To establish our main theorem, we need the following three lemmas. Lemma \ref{DS} and \ref{VC} is trivial to show by \eqref{VHK} if $f$ is smooth enough. The complete proofs of Lemma \ref{DS} and \ref{VC} can be found in \cite{Owen2005}; Lemma \ref{UB} is quite technical and proved in \cite{Heinrich2000}.
  \begin{lem}
    Let $f$ be defined on the hyper-rectangle $[a, b]$. Let $\{[a_i, b_i]: 1 \leq i\leq m < \infty\}$ be a split of $[a, b]$. Then
    \small\[\sum_{i = 1}^mV_{HK}^{[a_i, b_i]}(f) = V_{HK}^{[a, b]}(f)\]\normalsize
    \label{DS}
  \end{lem}
  \begin{lem}
    Let $f$ be defined on the hyper-rectangle $[a, b]$. Let $\tilde{f}(\tilde{x})$ be defined on the hyper-rectangle $[\tilde{a}, \tilde{b}]$ by $\tilde{f}(\tilde{x}) = f(x)$ where $x_i = \phi_i(\tilde{x}_i)$ with $\phi_i$ is a strictly monotone (increasing or decreasing) invertible function from $[\tilde{a}_i, \tilde{b}_i]$ onto $[a_i, b_i]$, then
    \small
    \[V_{HK}^{[\tilde{a}, \tilde{b}]}(\tilde{f}) = V_{HK}^{[a, b]}(f)\]
    \normalsize
    \label{VC}
  \end{lem}
  \begin{lem}
    Let $D_{n, d}^* = \inf_{x_1, ..., x_n\in [0, 1]^d}D_{n}^*(x_1, ..., x_n)$, we have \small\[D_{n, d}^* \leq cd^{1/2}n^{-1/2}\]\normalsize for all $n, d = 1, 2, ...,$ with a multiplicative constant $c$.
    \label{UB}
  \end{lem}
  \begin{rem}
    It is also shown that $c\leq 10$ in \cite{Aistleitner2011}. The asymptotic behavior of the star discrepancy on $n$ is much better (e.g., \emph{Halton sequence} \cite{Owen2003} has $D_n^* = O((\log n)^d/n)$); but it does not necessarily mean that the uniform bound which is valid for all $d$ and $n$ cannot be of order $n^{-1/2}$.
  \end{rem}
  \begin{thm}
    $f$ is defined on $d-$dimensional hyper-rectangle $[a, b]$ and $P = \{x_1, ..., x_n \in [a, b]\}$. Then we have
    \small
    \begin{equation}
    \Big|\int_{[a, b]}f(x)dx - \frac{\prod_{i = 1}^d(b_i - a_i)}{n}\sum_{i = 1}^nf(x_i)\Big|\leq \prod_{i = 1}^d(b_i - a_i)D_{n}^*(\tilde{P})V_{HK}^{[a, b]}(f)
    \label{thm2}
    \end{equation}
    \normalsize
    where $\tilde{P} = \{\tilde{x}_i = (\frac{x_{i1} - a_{1}}{b_{1}}, ..., \frac{x_{id} - a_{d}}{b_{d}})\}_{i = 1}^n$
    \label{theorem2}
  \end{thm}
  We are ready to state our main theorem,
  \begin{thm}
    $f$ is defined on hyper-rectangle $[0, 1]^d$ with $V_{HK}^{[0, 1]^d}(f)<\infty$ and the sub-rectangles $\{[a_i, b_i]\}_{i = 1}^l$ are a split of $[0, 1]^d$. Let $x_1, ..., x_N\in [0, 1]^d$ be an i.i.d sample set drawn from distribution $p(x)$ defined on $[0, 1]^d$ and $P_i = \{x_{i1}, ..., x_{in_i}, n_i\in\mathbb{N}^+\}$ are points in each sub-region. Consider a piecewise constant density estimator
    \small
    \[\hat{p}(x) = \sum_{i = 1}^ld_i\mathbf{1}\{x\in [a_i, b_i]\}\]
    \normalsize
    where $d_i = (\prod_{j = 1}^d(b_{ij} - a_{ij}))^{-1}n_i / N$, i.e., the empirical probability. In each sub-region $[a_i, b_i]$, $P_i$ satisfies
    \small
    \begin{equation}
    D_{n_i}^*(\tilde{P}_i)\leq \alpha_i D_{n_i, d}^*
    \label{cond}
    \end{equation}
    \normalsize
    where $\alpha_i = \sqrt{\frac{N}{n_id}}\frac{\theta}{c}$
    and $\theta$ is a positive constant; $\tilde{P}_i$ is defined as $\{\tilde{x}_j = (\frac{x_{j1} - a_{i1}}{b_{i1}}, ..., \frac{x_{jd} - a_{id}}{b_{id}})\}_{j = 1}^{n_i}$.
    Then
    \small
    \begin{equation}
    \Big|\int_{[0, 1]^d}f(x)\hat{p}(x)dx - \frac{1}{N}\sum_{i = 1}^Nf(x_i)\Big|\leq \frac{\theta}{\sqrt{N}} V_{HK}^{[0, 1]^d}(f)
    \label{converge}
    \end{equation}
    \normalsize
    \label{theorem3}
  \end{thm}
  \begin{rem}
    $\alpha_i$ controls the ``relative'' uniformity of the points and is adapted for $P_i$, i.e., it imposes more restricted constraint on the region containing large proportion of the sample ($n_i / N$).
  \end{rem}
  \begin{rem}
  In Monte Carlo methods, the convergence rate of $\frac{1}{N}\sum_{i = 1}^Nf(x_i)$ to $\int_{[0, 1]^d}f(x)p(x)dx$ is of order $O(1/\sqrt{N})$ as long as variance of $f(x)$ under $p(x)$ is bounded; our density estimate is optimal in the sense that it achieves the same rate of convergence.
  \end{rem}
  \begin{rem}
  There are many other $\hat{p}(x)$ satisfying \eqref{converge} such as the empirical distribution in the extreme or kernel density estimation with sufficiently small bandwidth. Our density estimation is attractive in the sense that it provides a very sparse summary of the data but still captures the landscape of the underlying distribution; moreover, the piecewise constant function does not suffer from many local bumps as kernel density estimation does.
  \end{rem}
  \begin{cor}
    For any hyper-rectangle $A = [a, b]\subset (0, 1)^d$. Let $\hat{P}(A) = \int_{A}\hat{p}(x)dx$ and $P(A) = \int_{A}p(x)dx$, then $|\hat{P}(A) - P(A)|$ converges to 0 at order $O(1/\sqrt{N})$ uniformly.
    \label{cor}
  \end{cor}
  \begin{rem}
    The total variation distance between probability measures $\hat{P}$ and $P$ is defined via
    $\delta(\hat{P}, P) = \sup_{A\in\mathcal{B}}|\hat{P}(A) - P(A)|$, where $\mathcal{B}$ is the Borel $\sigma$ algebra of $[0, 1]^d$; in contrast, Corollary \ref{cor} restricts $A$ to be rectangles.
  \end{rem}
%
%
  \section{Computational Aspects}
  There are no explicit formulas for calculating $D_{n}^*(x_1, ..., x_n)$ and $D_{n, d}^*$ except for low dimensions. If we replace $\alpha_i$ in \eqref{cond} and apply Lemma \ref{UB}, we actually intend to control $D_{n}^*(\tilde{P}_i)$ by $\theta\sqrt{N}/n_i$. There are several ways to approximate $D_{n}^*(x_1, ..., x_n)$: 1) E. Thi\'{e}mard presents an algorithm to compute the star discrepancy within a user specified error by partitioning the unit cube into subintervals \cite{Thiemard2000, Thiemard2001, Gnewuch2008}; 2) A genetic algorithm to calculate the lower bounds is proposed in \cite{Shah2010} and is used in our experiments; 3) A new randomized algorithm based on threshold accepting is developed in \cite{Gnewuch2012}. Comprehensive numerical tests indicate that it improves on other algorithms, especially in higher dimension $20\leq d\leq 50$. The interested readers are referred to the original articles for implementation details.

  In dealing with large data, several simple observations can be exploited to save computation: 1) it is trivial that $\max_{j = 1, ..., d} D_{n}^*(\{x_{ij}\}_{i = 1}^n)\leq D_{n}^*(\{x_i\}_{i = 1}^n)$. Let $x_{(i)j}$ be the $i$th smallest element in $\{x_{ij}\}_{i = 1}^n$, then $D_{n}^*(\{x_{ij}\}_{i = 1}^n) = \frac{1}{2n} + \max_{i = 1}^n|x_{(i)j} - \frac{2i - 1}{2n}|$ \cite{Doerr2013}, which has complexity $O(n\log n)$. Hence $\max_{j = 1, ..., d} D_{n}^*(\{x_{ij}\}_{i = 1}^n)$ can be used to compare against $\theta\sqrt{N}/n$ first before calculating $D_{n}^*(\{x_i\}_{i = 1}^n)$; 2) $\theta\sqrt{N}/n$ is large when $n$ is small, but $D_{n}^*(\{x_i\}_{i = 1}^n)$ is bounded above by 1; 3) $\theta\sqrt{N}/n$ is tiny when $n$ is large and $D_{n}^*(\{x_i\}_{i = 1}^n)$ is bounded below by $c_d\log^{(d - 1)/2}n^{-1}$ with some constant $c_d$ depending on $d$ \cite{Gnewuch2012a}; thus we can keep splitting without checking \eqref{cond} when $\theta\sqrt{N}/n\leq\epsilon$, where $\epsilon$ is a small positive constant (say 0.001) specified by the user. This strategy may introduce few more sub-rectangles, but the running time gain is considerable.

  Another approximation works well in practice is by replacing star discrepancy with computationally attractive $\mathcal{L}_2$ star discrepancy, i.e., $D_{n}^{(2)}(x_1, ..., x_n) = (\int_{[0, 1]^d}|\frac{1}{n}\sum_{i = 1}^n\mathbf{1}_{x_i\in[0, a)} - \prod_{i = 1}^da_i|^2da)^{1/2}$; in fact, several statistics to test uniformity hypothesis based on $D_{n}^{(2)}$ are proposed in \cite{Liang2001}; however, the theoretical guarantee in Theorem \ref{theorem3} is no longer valid. By Warnock's formula \cite{Doerr2013},
  \small
  \[[D_{n}^{(2)}(x_1, ..., x_n)]^2 = \frac{1}{3^d} - \frac{2^{1 - d}}{n}\sum_{i = 1}^n\prod_{k = 1}^d(1 - x_{ik}^2) + \frac{1}{n^2}\sum_{i, j = 1}^n\prod_{k = 1}^d\min\{1 - x_{ik}, 1 - x_{jk}\}\]
  \normalsize
  $D_{n}^{(2)}$ can be computed in $O(n\log^{d - 1}n)$ by K. Frank and S. Heinrich's algorithm \cite{Doerr2013}. At each scan of $\mathcal{B}$ in Algorithm \ref{algo}, the total complexity is at most $\sum_{i = 1}^lO(n_i\log^{d - 1}n_i)\leq\sum_{i = 1}^lO(n_i\log^{d - 1}n)\leq O(n\log^{d - 1}n)$.
  \section{Experimental Results}
  \subsection{Simulations and Comparisons}

\begin{figure}[ht]
    \center
    \includegraphics[width=.7\textwidth]{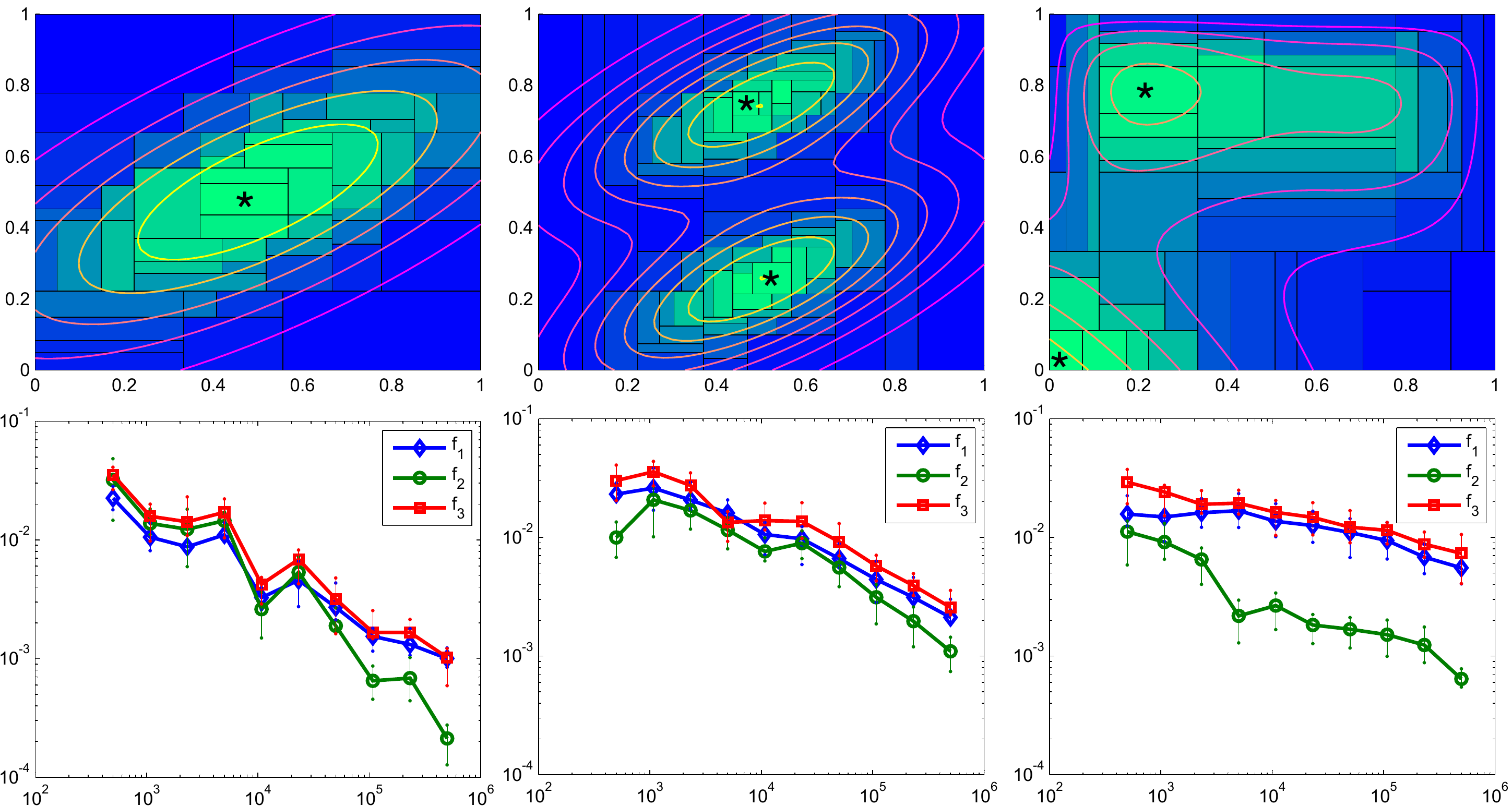}
    \caption{First row: simulation on 2D data with 3 different densities; the modes are marked by stars. Second row: simulation on 2, 5 and 10 dimensional data (from left to right) with sample functions $f_1, f_2, f_3$.}
    \label{simulation}
  \end{figure}

    \textbf{1)} To demonstrate the methods and visualize the results, we simulate our algorithm through 3 2-dimensional data sets generated from 3 distributions respectively, i.e., $x\sim \mathcal{N}(\mu, \Sigma)\mathbf{1}\{x\in [0, 1]^2\}$ with $\mu = (.50, .50)^T, \Sigma = [0.08, 0.02;0.02, 0.02]$; $x\sim (1/2)\mathcal{N}(\mu_1, \Sigma_1)\mathbf{1}\{x\in [0, 1]^2\} + (1/2)\mathcal{N}(\mu_2, \Sigma_2)\mathbf{1}\{x\in [0, 1]^2\}$, with $\mu_1 = (.50, .25)^T, \Sigma_1 = [0.04, 0.01;0.01, 0.01]$ and $\mu_2 = (.50, .75)^T, \Sigma_2 = [0.04, 0.01;0.01, 0.01]$; $x\sim (1/3)\beta(2, 5)\beta(5, 2) + (1/3)\beta(4, 2)\beta(2, 4) + (1/3)\beta(1, 3)\beta(3, 1)$; where $\mathcal{N}$ is the Gaussian distribution and $\beta$ is the beta distribution. The size of each data set is 10,000. As shown in the first row of Figure \ref{simulation}, we draw the partitions on 2D and render the estimated densities with a color map; the corresponding contours of true densities are embedded for comparison purpose.\\
    \textbf{2)} To evaluate the theoretical bound \eqref{converge}, we choose 3 simple reference functions with dimension $d = 2$, 5 and 10 respectively, i.e., $f_1(x) = \sum_{i = 1}^n\sum_{j = 1}^dx_{ij}^{1/2}$, $f_2(x) = \sum_{i = 1}^n\sum_{j = 1}^dx_{ij}$, $f_3(x) = (\sum_{i = 1}^n\sum_{j = 1}^dx_{ij}^{1/2})^2$ and samples are generated from $p(x) = \frac{1}{2}\Big(\prod_{i = 1}^d\beta_{15, 5}(x_i) + \prod_{i = 1}^d\beta_{5, 15}(x_i)\Big)$. The error $|\int_{[0, 1]^d}f_i(x)p(x)dx - \int_{[0, 1]^d}f_i(x)\hat{p}(x)dx|$ is bounded by
  $|\int_{[0, 1]^d}f_i(x)p(x)dx - \frac{1}{n}\sum_{j = 1}^nf_i(x_j)| + |\int_{[0, 1]^d}f_i(x)\hat{p}(x)dx - \frac{1}{n}\sum_{j = 1}^nf_i(x_j)|$
  where $\hat{p}(x)$ is the estimated density; the first term is controlled by $O(n^{-1/2})$ which is well known in Monte Carlo methods and the second term is controlled by \eqref{converge}, thus the error is of order $O(n^{-1/2})$. By varying the data size, the relative error vs. sample size is plotted on log-log scale for each dimension in the second row of Figure \ref{simulation}, their standard errors are obtained through generating 10 replicas under same distributions. Interestingly, the linear pattern shows up as expected.

  \textbf{3)} There are other two classes of domain partition based density estimators: I) Optional P\'{o}lya Tree (OPT) \cite{Wong2010} is a class of conjugate nonparametric priors based on binary partition and optional stopping; II) Bayesian Sequential Partitioning (BSP) \cite{Lu2013} is introduced as a computationally more attractive alternative to OPT and simulations show that the density constructed by BSP is very close to MAP of OPT. However, the density estimate of OPT or BSP is obtained by sampling the posterior and the finite-sample error bound is not provided, while ours is constructed from a frequentist perspective and we establish a theoretical framework for error analysis. The recursion in OPT has exponential complexity and BSP in principle searches among the exponential number of possible partitions, whereas our partitioning scheme is greedy and results in significant speedup. As to binary partition, we no longer restrict the algorithm to split the hyper-rectangle evenly (in the middle); by introducing the ``gap'', we do the partitioning more adaptive to the data.

   To show the efficiency and scalability of our method, we compare it with KDE, OPT and BSP in terms of estimation error and running time. We simulate samples from $x\sim (\sum_{i = 1}^4\pi_i \mathcal{N}_i(\mu_i, \Sigma))\mathbf{1}\{x\in [0, 1]^d\}$ with $d = \{2, 3, 4, 5, 6\}$ and $n = \{10^3, 10^4, 10^5\}$ respectively (refer to Appendix A for detailed experiment settings).

  The estimation error and running time are summarized in Table \ref{tab1} and Table \ref{tab2} respectively, the standard error is obtained by generating 20 replicas for each $(d, n)$ pair. 

  \begin{table*}[ht]
  \centering
  \scalebox{0.7}{
    \begin{tabular}{@{}r|rrrr|rrrr|rrrr|@{}}
    \toprule
  & \multicolumn{4}{c}{Error($n = 10^3$)}&\multicolumn{4}{c}{Error($n = 10^4$)} & \multicolumn{4}{c}{Error($n = 10^5$)}\\
  \hline
  & KDE & OPT & BSP & ours & KDE & OPT & BSP & ours & KDE & OPT & BSP & ours\\
  \midrule
  $d$\\
  \hline
  $2$  &\textbf{0.2331}   &0.2442   &0.2533   &0.2634     &0.1604   &0.1637    &0.1622   &\textbf{0.1603}   &\textbf{0.0305}    &0.0376   &0.0308   &0.0312    \\
       &(0.0221) &(0.0211) &(0.0163) &(0.0207)   &(0.0102) &(0.0101)  &(0.0113) &(0.0132) &(0.0021)  &(0.0021)  &(0.0025) &(0.0027)    \\
       \hline
  $3$  &0.2893   &0.2751   &0.2683   &\textbf{0.2672}     &0.2163   &0.1722    &\textbf{0.1717}   &0.1721   &0.0866    &0.0460   &0.0477   &\textbf{0.0452}      \\
       &(0.0227) &(0.0212) &(0.0233) &(0.0265)   &(0.0199) &(0.0222)  &(0.0183) &(0.0143) &(0.0047)  &(0.0049) &(0.0043) &(0.0034)    \\
       \hline
  $4$  &0.3313   &\textbf{0.2807}   &0.2872   &0.2855     &0.2466   &\textbf{0.1832}    &0.1882   &0.1955   &0.1600    &*        &\textbf{0.0629}   &0.0637      \\
       &(0.0225) &(0.0232) &(0.0217) &(0.0191)   &(0.0113) &(0.0185)  &(0.0147) &(0.0185) &(0.0057)  &*        &(0.0061) &(0.0059)    \\
       \hline
  $5$  &0.3522   &0.3001   &0.3035   &\textbf{0.2907}     &0.2599   &\textbf{0.1911}    &0.1987   &0.1963   &0.1817    &*        &\textbf{0.0716}   &0.0721      \\
       &(0.0317) &(0.0299) &(0.0319) &(0.0302)   &(0.0199) &(0.0143)  &(0.0122) &(0.0131) &(0.0088)  &*        &(0.031)  &(0.0066)    \\
       \hline
  $6$  &0.4011   &\textbf{0.3512}   &0.3515   &0.3527     &0.2833   &*         &0.2093   &\textbf{0.2011}   &0.1697    &*        & *       &\textbf{0.0809}      \\
       &(0.0318) &(0.0307) &(0.0354) &(0.381)    &(0.0255) &*         &(0.0166) &(0.0137) &(0.0122)  &*        & *       &(0.0071)    \\
  \bottomrule
  \end{tabular}
  }
  \caption{The \textbf{error in Hellinger Distance} between KDE, OPT, BSP, our method and the true density for each pair $(d, n)$ respectively. Stars indicate that the running time exceeds 3600s. The numbers in parentheses are standard errors of 20 replicas.}
  \label{tab1}
  \end{table*}

  \begin{table*}[ht]
  \centering
  \scalebox{0.7}{
    \begin{tabular}{@{}r|rrrrrrrrrrrr@{}}
    \toprule
  & \multicolumn{3}{c}{Runing Time($n = 10^3$)}&&\multicolumn{3}{c}{Runing Time($n = 10^4$)} && \multicolumn{3}{c}{Runing Time($n = 10^5$)}\\
  \cline{2-4}  \cline{6-8} \cline{10-12}
  & OPT & BSP & ours && OPT & BSP & ours && OPT & BSP & ours\\
  \midrule
  $d$\\
  \hline
  $2$  & 0.4(0.0) & 1.2(0.1) & 1.7(0.1) && 2.8(0.1) & 23.2(6.4) & 11.2(0.9) && 42.9(0.3) & 263.1(44.9) & 95.8(3.6)\\
       \hline
  $3$  & 0.8(0.0) & 1.6(0.3) & 2.2(0.4) && 13.3(1.1) & 27.7(8.4) & 17.1(1.9) && 252(2.8) & 422.8(91.7) & 143.7(4.3)\\
       \hline
  $4$  & 1.7(0.1) & 3.5(0.2) & 3.3(0.8) && 137.7(10.2) & 42.3(5.3) & 22.6(2.0) && * & 684.3(80.2) & 192.4(5.1)\\
       \hline
  $5$  & 75.6(3.3) & 4.9(0.3) & 3.2(0.7) && 1731.7(17.7) & 138.2(9.7) & 21.3(2.2) && * & 1547.9(155.6) & 231.6(6.8)\\
       \hline
  $6$  & 251.3(7.9) & 5.1(0.4) & 3.8(0.7) && * & 179.1(13.4) & 30.0(2.1) && * & * & 285.4(10.2)\\
  \bottomrule
  \end{tabular}
  }
  \caption{The \textbf{average running time in seconds} of OPT, BSP and our method for each pair $(d, n)$ respectively. Stars indicate that the running time exceeds 3600s. The numbers in parentheses are standard errors of 20 replicas. OPT and BSP are implemented in \texttt{C++} and our method is in \texttt{Matlab}; it is noticed that the latency of \texttt{Matlab} dominates in small simulations.}
  \label{tab2}
  \end{table*}

  \subsection{Mode Detection}
  A direct application of the piecewise constant density is to detect modes \cite{Comaniciu2002}, i.e., the dense areas or local maxima on the domain. The modes of our density estimator is defined as
  \begin{defn}
    A mode of the piecewise constant density is a sub-rectangle in the partition that its density is largest among all its neighbors as indicated by the stars in Figure \ref{simulation}.
    \label{mode_def}
  \end{defn}
  \begin{figure}[ht]
    \center
    \includegraphics[width = 1.0\textwidth]{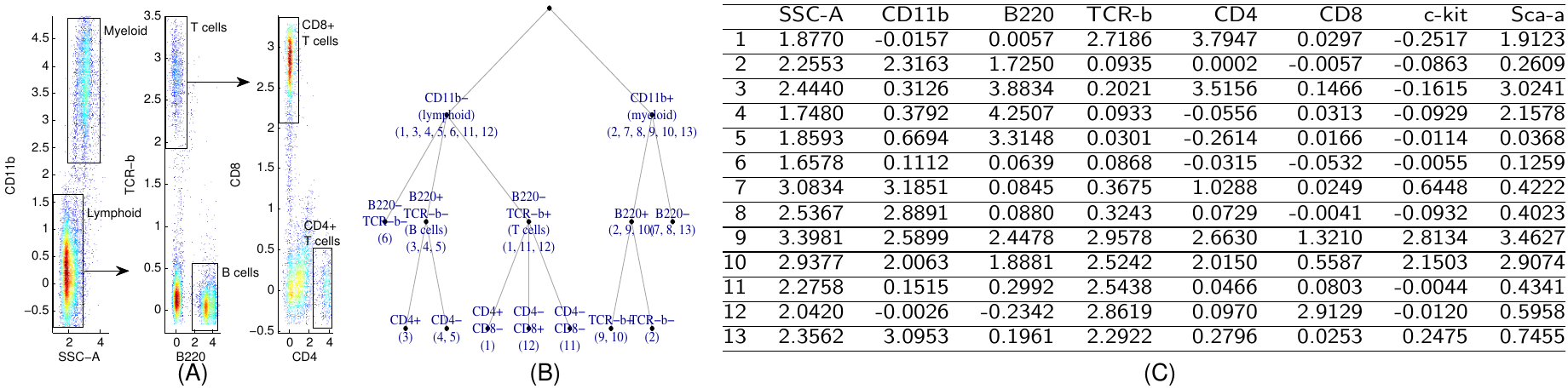}
    \caption{\textbf{Flow Cytometry.} (A): an illustrative gating sequence, the cell type in each gate is attached; (B) there are 13 modes detected by our algorithm, and we arrange these modes into a hierarchical dendrogram: at first level, they are grouped by expression levels of CD11b; subsequently, the CD11b- modes are grouped according to B220 and TCR-b then further splitted according to CD4 and CD8 on the next level; the CD11b+ modes are grouped by B220 then by TCR-b; (C) the details of the expression levels of each mode.}
    \label{mouse}
  \end{figure}

  Flow cytometry allows measuring simultaneously multiple characteristics of a large number of cells and is a ubiquitous and indispensable technology in medical settings. One effort of current research is to identify homogeneous sub-populations of cells automatically instead of manual gating, which is criticized for its subjectivity and non-scalability. There are a large amount of recent literatures concerning on auto gating and clustering, see \cite{Aghaeepour2013} and many reference therein.

  In order to apply our method, we regard each cell as one observation in the sample space, i.e., if there are $n$ markers attached to a single cell, then the whole data set is generated from a hypothetical $n$ dimensional distribution. Mature cell populations concentrate in some high density areas, which can be easily identified in the binary partitioned space by Definition \ref{mode_def}.

  One practical issue needs to be addressed for most of the Cytometry analysis techniques: there is asymmetry in sub-populations; by optimizing a predefined loss function, it is possible that some sparse yet crucial populations are overlooked if the algorithms take most of the efforts to control the loss in denser areas. A remedy for this issue is to perform a down-sampling \cite{Aghaeepour2013, Qiu2011} step to roughly equalize the densities among populations then up-sampling after populations are identified; however, this step is dangerous that it may fails to sample enough cells in sparse populations, as a result, these populations are lost in the down-sampled data. In contrast, our approach does not require down-sampling step, and the asymmetry among populations is captured by their densities.

  For the mouse bone marrow data studied in \cite{Qiu2011}, we choose the 8 markers (SSA-C, CD11b, B220, TCR-$\beta$, CD4, CD8, c-kit, Sca-1) that are relevant to the cell types of interests; the number of cells is $\thicksim$380,000 after removing mutli-cell aggregates and co-incident events. 13 sub-populations are identified by our algorithm (\cite{Qiu2011} and its supplementary materials), the results are summarized in Figure \ref{mouse}.
  \subsection{Density Topology Exploration and Visualization}\label{soa}
  Level set tree \cite{Zhou2009} a.k.a. connectivity graph (DG), is widely used to represent energy landscapes of systems. It summarizes the hierarchy among various local maxima and minima in the configuration space. Its topology is a tree and each inner node on the tree is a changing point that merges two or more independent regions in the domain. With the density estimation at hand, one may construct DG for samples instead of a given energy or density function. Unlike KDE that suffers from many local bumps and results in an overly complicated DG, $\eqref{eq1}$ is well suited for this purpose, partially because it smoothes out the minor fluctuations and takes only limited number of values; moreover, the simple structure of \eqref{eq1} makes the construction of such graph easy (i.e., one can just scan through each $r_i$ in decreasing order of $d(r_i)$, the complete algorithm is given in Appendix \ref{LST}). The DG of \eqref{eq1} not only reveals the modes of the density on its leaves, it also provides a tool to visualize high dimensional data hierarchically; for example, in fiber tractography \cite{Kent2013}, DG is used to visualize and analyze topography in fiber streamlines interactively.

\begin{figure}[ht]
    \center
    \includegraphics[width = .81\textwidth]{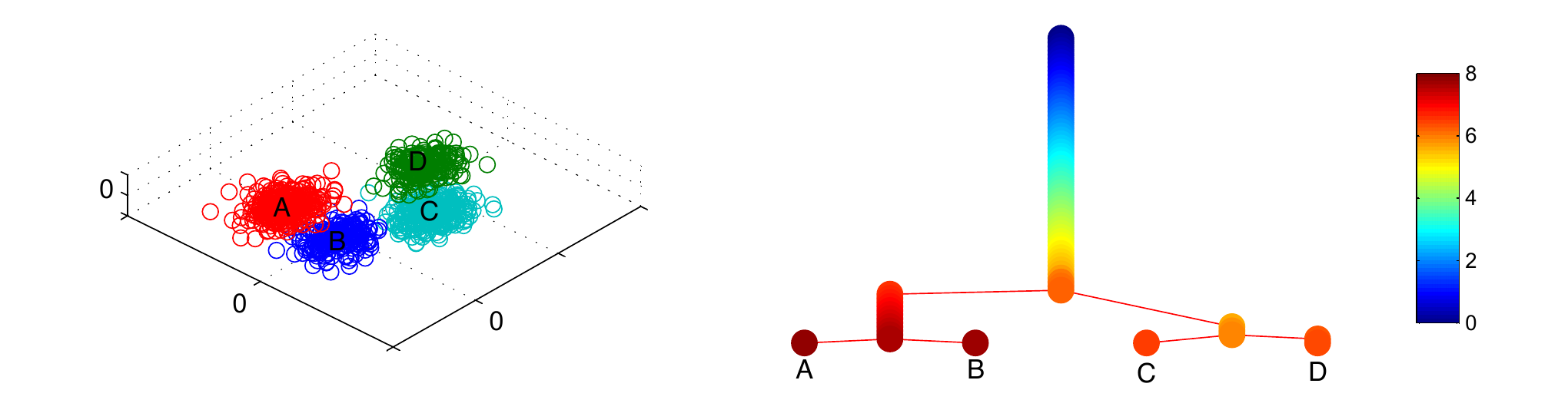}
    \caption{\textbf{Level Set Tree.} Left: the samples are generated from a Gaussian Mixture with 4 modes. Right: the level set tree. The clusters are annotated by A, B, C, D and colors are the levels.}
    \label{LS}
  \end{figure}

  We demonstrate that how our piecewise density function can be used to construct level set trees in Figure \ref{LS}. The basic pipeline is to scan sub-rectangles sequentially according to the decreasing order of their densities and agglomerate the sub-rectangles according to their adjacency.

  \section{Conclusion and Future Work}
  We have developed a nonparametric density estimation framework based on discrepancy criteria, proved its theoretical properties and shown that it is applicable to different types of problems. We point out several future research directions of interest: 1) Current approach deals with continuous features, but how to extend our theories and algorithm to handle (unordered) categorical data? 2) Coordinate-wise partition limits the approximation capability, recent progress \cite{Basu2014} in Quasi Monte Carlo on simplex provides us a possible alternative to use more flexible partition schemes. 3) Theoretically, how to sharpen Corollary \ref{cor} in order to enlarge the class of Borel sets rather than rectangles? 4) A thorough comparison is necessary to understand the empirical differences between our method and OPT or BSP. 5) Our approach can be viewed as an unsupervised version of CART (Classification And Regression Trees), therefore, it would be interesting to explore the idea of ensemble methods in analogous to boosting and random forest \cite{Criminisi2011}.

  \bibliographystyle{abbrv}
  \bibliography{nips}
\newpage
\appendix
\section{Proofs}
\subsection{Proof of Theorem \ref{theorem2}}
  \begin{proof}
  Define $\tilde{f}(\tilde{x}) = f(x)$, where $\tilde{x} = (\frac{x_{1} - a_{1}}{b_{1}}, ..., \frac{x_{d} - a_{d}}{b_{d}})$ and apply Theorem \ref{KH} to $\tilde{f}(\tilde{x})$, we have
  \[\Big|\int_{[0, 1]^d}\tilde{f}(\tilde{x})d\tilde{x} - \frac{1}{n}\sum_{i = 1}^n\tilde{f}(\tilde{x}_i)\Big|\leq D_{n}^*(\tilde{P})V_{HK}^{[0, 1]^d}(\tilde{f})\]
  From Lemma \ref{VC}, $V_{HK}^{[0, 1]^d}(\tilde{f}) = V_{HK}^{[a, b]}(f)$; $\int_{[0, 1]^d}\tilde{f}(\tilde{x})d\tilde{x} = (\prod_{i = 1}^d(b_i - a_i))^{-1}\int_{[a, b]}f(x)dx$ by change of variables and $\tilde{f}(\tilde{x}_i) = f(x_i)$ by definition. Hence, \eqref{thm2} follows immediately.
  \end{proof}

\subsection{Proof of Theorem \ref{theorem3}}
  \begin{proof}
    Apply Theorem \ref{theorem2} to each $[a_i, b_i], i = 1, ..., l$, we have
    \begin{equation}
      \Big|\int_{[a_i, b_i]}f(x)dx - \frac{\prod_{j = 1}^d(b_{ij} - a_{ij})}{n_i}\sum_{j = 1}^{n_i}f(x_{ij})\Big|\leq \prod_{j = 1}^d(b_{ij} - a_{ij})D_{n_i}^*(\tilde{P_i})V_{HK}^{[a_i, b_i]}(f)
      \label{eqUB}
    \end{equation}
    and by triangular inequality, we have
    \begin{equation}
      \Big|\int_{[0, 1]^d}f(x)\hat{p}(x)dx - \frac{1}{N}\sum_{i = 1}^Nf(x_i)\Big|\leq \sum_{i = 1}^ld_i\Big|\int_{[a_i, b_i]}f(x)dx - \frac{1}{d_iN}\sum_{j = 1}^{n_i}f(x_{ij})\Big|
    \end{equation}
    By the definition of $d_i$, $d_iN = (\prod_{j = 1}^d(b_{ij} - a_{ij}))^{-1}n_i$; combine with Theorem \ref{theorem2}, \eqref{cond}, \eqref{eqUB} and Lemma \ref{UB}, we have
    \begin{eqnarray*}
      \sum_{i = 1}^ld_i\Big|\int_{[a_i, b_i]}f(x)dx - \frac{1}{d_iN}\sum_{j = 1}^{n_i}f(x_{ij})\Big|\leq \sum_{i = 1}^ld_i\prod_{j = 1}^d(b_{ij} - a_{ij})D_{n_i}^*(\tilde{P_i})V_{HK}^{[a_i, b_i]}(f)\\
      \leq \sum_{i = 1}^l\frac{n_i}{N}\sqrt{\frac{N}{n_id}}\frac{\theta}{c}D_{n_i, d}^*V_{HK}^{[a_i, b_i]}(f)\leq
      \sum_{i = 1}^l\frac{n_i}{N}\sqrt{\frac{N}{n_id}}\frac{\theta}{c}cd^{1/2}n_i^{-1/2}V_{HK}^{[a_i, b_i]}(f)\quad\quad\quad\quad\\
      =\frac{\theta}{\sqrt{N}} \sum_{i = 1}^lV_{HK}^{[a_i, b_i]}(f)=\frac{\theta}{\sqrt{N}} V_{HK}^{[0, 1]^d}(f)\quad\quad\quad\quad\quad\quad\ \ \quad\quad\quad\quad\textrm{Apply Lemma \ref{DS}}
    \end{eqnarray*}
  \end{proof}

\subsection{Proof of Corollary \ref{cor}}
  \begin{proof}
     In Monte Carlo methods, the convergence rate of $\frac{1}{N}\sum_{i = 1}^Nf(x_i)$ is of order $O(\frac{\textrm{std}(f)}{\sqrt{N}})$. Let $f(x) = \mathbf{I}\{x\in [a, b]\} = \mathbf{I}_{[a, b]}$ be defined on $[0, 1]^d$, we have $\textrm{var}(f) = P(A)(1 - P(A))\leq 1/4$; thus, this error is bounded uniformly.

     If another indicator function $\tilde{f}$ is defined on $[\tilde{a}, \tilde{b}]\subset (0, 1)^d$, then let
     \[\phi_j(\tilde{x}_j) = \frac{a_j}{\tilde{a}_j}\tilde{x}_j\mathbf{I}_{[0, \tilde{a}_j)} + (a_j + \frac{b_j - a_j}{\tilde{b}_j - \tilde{a}_j}(\tilde{x}_j - \tilde{a}_j))\mathbf{I}_{[\tilde{a}_j, \tilde{b}_j)} + (b_j + \frac{1 - b_j}{1 - \tilde{b}_j}(\tilde{x}_j - \tilde{b}_j))\mathbf{I}_{[\tilde{b}_j, 1]}\]
     and $\phi(\tilde{x}) = \prod_{j = 1}^d\phi_j(\tilde{x}_j)$ and apply Lemma \ref{VC}, we have $V_{HK}^{[0, 1]^d}(\tilde{f}) = V_{HK}^{[0, 1]^d}(f)$; thus, the left term of \eqref{converge} is bounded uniformly.

     The error $|\int_{[0, 1]^d}f_i(x)p(x)dx - \int_{[0, 1]^d}f_i(x)\hat{p}(x)dx|$ is bounded by
  $|\int_{[0, 1]^d}f_i(x)p(x)dx - \frac{1}{n}\sum_{j = 1}^nf_i(x_j)| + |\int_{[0, 1]^d}f_i(x)\hat{p}(x)dx - \frac{1}{n}\sum_{j = 1}^nf_i(x_j)|$. Combining the two parts, the theorem follows by triangular inequality.
  \end{proof}
\section{Experiment Setting of Comparison with OPT and BSP}
  The source codes are obtained from the authors. Their implementation language is C++; in contrast, our method is implemented in Matlab. For small data, the latency of Matlab dominates the computing time as shown in the first block of Table \ref{tab2}.

  \begin{equation*}
  \left(\begin{array}{c}
  \mu_1\\
  \mu_2\\
  \mu_3\\
  \mu_4
  \end{array}\right) = \left(\begin{array}{rrrrr}
  1/4 & 1/4 & 1/2 & \cdots & 1/2\\
  1/4 & 3/4 & 1/2 & \cdots & 1/2\\
  3/4 & 1/4 & 1/2 & \cdots & 1/2\\
  3/4 & 3/4 & 1/2 & \cdots & 1/2\\
  \end{array}\right)_{4\times d}
  \end{equation*}
  and $\Sigma = 0.01\mathbf{I}$, i.e., the identity matrix, $\pi = (1/4, 1/4, 1/4, 1/4)$. The system where the comparison is performed is Ubuntu 13.04, AMD64
8-core Opteron 2384 (SB X6240, 0916FM400J); 31.42GB RAM, 10GB swap.

\section{Level Set Tree Algorithm}\label{LST}
For a given partition $P$ and the list of pairs of sub-regions and corresponding densities $\{r_i, d(r_i)\}_{i = 1}^l$ as in \eqref{eq1}, we build a graph $G$ based on the adjacency of sub-regions and each sub-region is a node on the graph. The algorithm to determine the adjacency of sub-region $i, j$ is:
\begin{algorithmic}[1]
\Procedure{is-adjacent}{$r_i,r_j$}
\State $c_k = (c_{k1}, ..., c_{kd})$: the center of $r_k, k\in \{i, j\}$
\State $l_k = (l_{k1}, ..., l_{kd})$: the width of $r_k$ in each dimension, $k\in\{i, j\}$
\For{$k\gets 1, ..., d$}
    \If{$|c_{ik} - c_{jk}| > (l_{ik} + l_{jk}) / 2$}
        \State \textbf{return} False
    \EndIf
\EndFor
\State \textbf{return} True
\EndProcedure
\end{algorithmic}
$G$ is constructed by connecting adjacent sub-regions. When $G$ has $k > 1$ connected components $\{c_1, c_2, ..., c_k\}$, a ``virtual region'' $r$ is added into $G$ with density zero and it connects the region in each $\{c_i\}_{i = 1}^k$ with lowest density in order to make $G$ connected.

Starting with empty set $S_0$ at step 0, the sub-region is added into $S$ sequentially according to the decreasing order of densities. At $(i - 1)$th step, we compute the connected components based on the induced sub-graph $G(S_{i - 1})$. Suppose $S_{i - 1} = \{g_1, g_2, ..., g_j\}$, where $g_i$ is the connected components and $j$ is the number of components; at step $i$, there are two scenarios when $r$ is added into $S_{i - 1}$: i) $r$ is adjacent to multiple components $\{g_{i_1}, g_{i_2}, ..., g_{i_t}\}$, then $S_{i} = \{r_{(i)}\cup\{g_{i_j}\}_{j = 1}^t, S_{i - 1}\backslash \cup\{g_{i_j}\}_{j = 1}^t\}$ and $r$ is the parent of latest added sub-region in each $g_{i_j}$; ii) $r$ is disconnected with all the components, then $S_{i} = \{S_{i - 1}, r\}$ and $r$ is a leaf.
The complete description of the algorithm is:
\begin{algorithmic}[1]
\Procedure{sub-level-tree}{$P$}
\State Build $G$ from $P$
\State $S_0 = \emptyset$
\State $R_0 = \emptyset$
\For{$i\gets 1, ..., l$}\Comment $n$ is the number of nodes in $G$
    \State $S_{i-1} = \{g_1, g_2, ..., g_j\}$\Comment $\{g_k\}_{k = 1}^j$ are the connected components in $G(S_{i-1})$
    \State $R_{i-1} = \{r_1, r_2, ..., r_j\}$\Comment $r_k$ is the last sub-region added into $g_k, k = 1, ..., j$
    \If{$r_{(i)}$ is adjacent to $\{g_{i_1}, g_{i_2}, ..., g_{i_t}\}$}\Comment $r_{(i)}$ has $i$th largest density in $P$
    \State $S_{i} = \{r_{(i)}\cup\{g_{i_j}\}_{j = 1}^t, S_{i - 1}\backslash \cup\{g_{i_j}\}_{j = 1}^t\}$
    \State $R_{i} = \{r_{(i)}, R_{i - 1}\backslash\cup \{r_{i_j}\}_{j = 1}^t\}$
    \State $\wp(r_{i_j}) = r_{(i)}, j = 1, ..., t$\Comment $\wp$ stores the parent of each sub-region
    \State Color$(r_{(i)})$ = density$(r_{(i)})$
    \Else
    \State $S_{i} = \{S_{i - 1}, r_{(i)}\}$
    \State $R_{i} = \{R_{i - 1}, r_{(i)}\}$
    \State Color$(r_{(i)})$ = density$(r_{(i)})$
    \EndIf
\EndFor
\State \textbf{return} $\wp$, Color
\EndProcedure
\end{algorithmic}
\end{document}